\documentclass[sigconf]{acmart}
\usepackage{fancyvrb}
\usepackage{adjustbox}
\usepackage{multirow}
\usepackage{comment}
\usepackage{amsmath}
\AtBeginDocument{%
  \providecommand\BibTeX{{%
    \normalfont B\kern-0.5em{\scshape i\kern-0.25em b}\kern-0.8em\TeX}}}

\copyrightyear{2021}
\acmYear{2021}
\setcopyright{acmcopyright}\acmConference[KDD '21]{Proceedings of the 27th ACM SIGKDD Conference on Knowledge Discovery and Data Mining}{August 14--18, 2021}{Virtual Event, Singapore}
\acmBooktitle{Proceedings of the 27th ACM SIGKDD Conference on Knowledge Discovery and Data Mining (KDD '21), August 14--18, 2021, Virtual Event, Singapore}
\acmPrice{15.00}
\acmISBN{978-1-4503-8332-5/21/08}
\acmDOI{10.1145/3447548.3467164}

\usepackage{etoolbox}
\makeatletter
\patchcmd{\maketitle}
 {\def\@makefnmark}
 {\def\@makefnmark{}\def\useless@macro}
 {}{}
\makeatother

\begin{document}
\fancyhead{}
\title{PAM: Understanding Product Images \\ in Cross Product Category Attribute Extraction}

\author{Rongmei Lin$^{1\text{*}}$, Xiang He$^{2}$, Jie Feng$^{2}$, Nasser Zalmout$^{2}$, Yan Liang$^{2}$, Li Xiong$^{1}$, Xin Luna Dong$^{2}$}\titlenote{Work performed during internship at Amazon.}
\affiliation{
	\institution{$^1$Emory University  $\quad$ $^2$Amazon}
	\institution{$^1$\{rlin32,lxiong\}@emory.edu  $\quad$ $^2$\{xianghe,jiefeng,nzalmout,ynliang,lunadong\}@amazon.com }
}


\begin{abstract}

Understanding product attributes plays an important role in improving online shopping experience for customers and serves as an integral part for constructing a product knowledge graph. Most existing methods focus on attribute extraction from text description or utilize visual information from product images such as shape and color. Compared to the inputs considered in prior works, a product image in fact contains more information, represented by a rich mixture of words and visual clues with a layout carefully designed to impress customers. This work proposes a more inclusive framework that fully utilizes these different modalities for attribute extraction. Inspired by recent works in visual question answering, we use a transformer based sequence to sequence model to fuse representations of product text, Optical Character Recognition (OCR) tokens and visual objects detected in the product image. The framework is further extended with the capability to extract attribute value across multiple product categories with a single model, by training the decoder to predict both product category and attribute value and conditioning its output on product category. The model provides a unified attribute extraction solution desirable at an e-commerce platform that offers numerous product categories with a diverse body of product attributes.  We evaluated the model on two product attributes, one with many possible values and one with a small set of possible values, over 14 product categories and found the model could achieve 15\% gain on the Recall and 10\% gain on the F1 score compared to existing methods using text-only features.
\end{abstract}

\maketitle

\section{Introduction}

\begin{figure}[h]
  \centering
  \includegraphics[width=\linewidth]{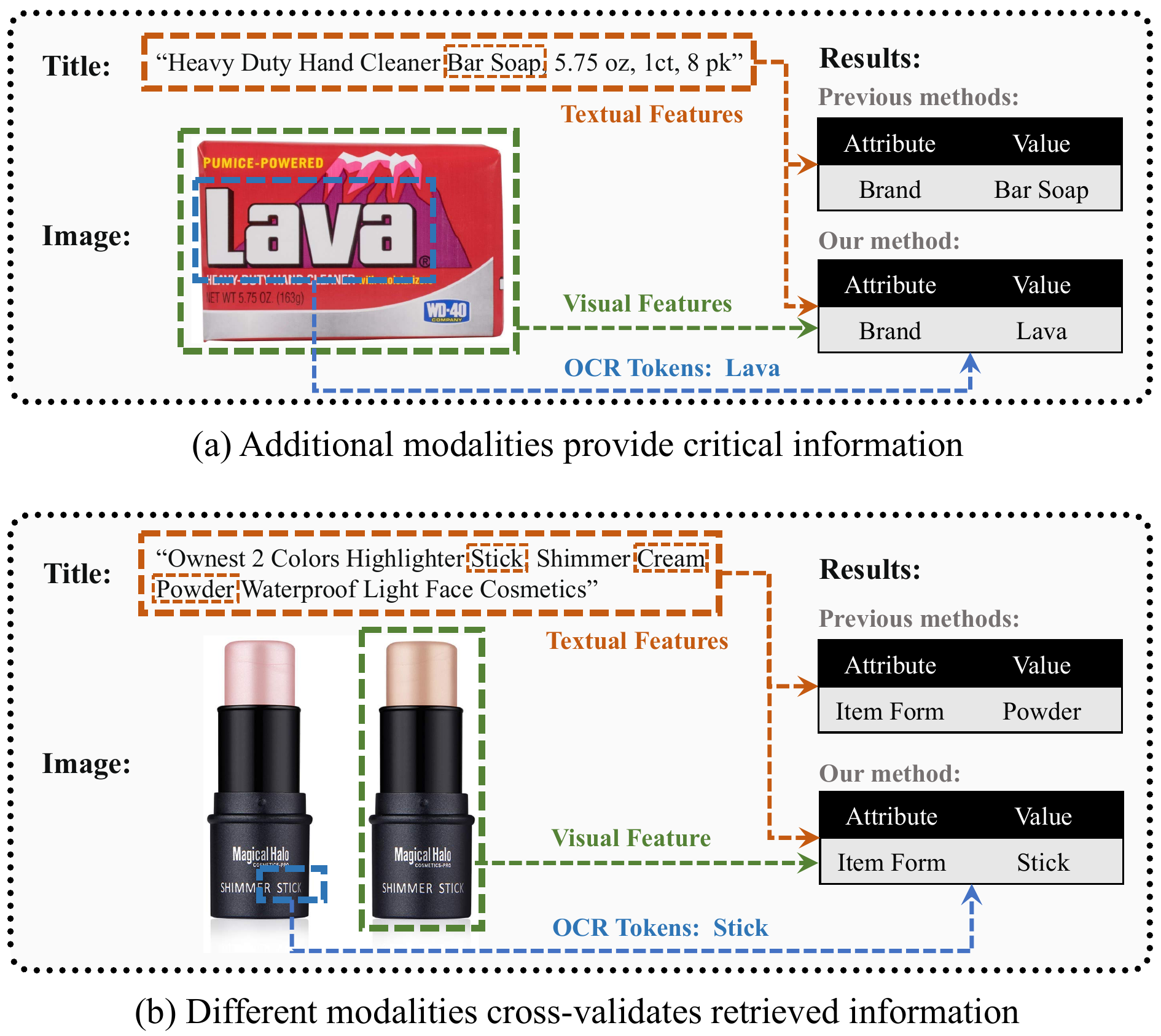}
  \caption{Compared to existing attribute value extraction works which focused on text-only input features, our approach is able to take extra multi-modal information (Visual Features and OCR Tokens) from product image as input.}
  \label{fig:sample}
\end{figure}

Product attributes, such as brand, color and size, are critical
product features that customers typically use to differentiate one product
from another. Detailed and accurate attribute values can make it easier for customers to
find the products that suit their needs and give them a
better online shopping experience. It may also increase the customer base and
revenue for e-commerce platforms. Therefore, accurate product attributes are 
essential for e-commerce applications such as product
search and recommendation.  Due to the manual and tedious 
nature of entering product information~\cite{dong2020autoknow}, the product attributes acquired from a
massive number of retailers and manufacturers on e-commerce platforms are usually incomplete, noisy and prone to errors. To address this challenge, there has been a surge of interest in automatically extracting attribute values from readily available product profiles~\cite{zheng2018opentag,xu2019scaling,karamanolakis2020txtract, wang2020learning, dong2020autoknow, zhang2021vinvl}. Most of the existing works rely only on the textural cues obtained from the text descriptions in the product profiles, which is often far from sufficient to capture the target attributes.

In this work, we propose to leverage the product images in the attribute value extraction task. Besides the commonly used textural features from product descriptions, our approach simultaneously utilizes the generic visual features and the textual content hidden in the images, which are extracted through Optical Character Recognition (OCR). We studied 30 popular product attributes applicable to different product categories, including electronics, home innovation, clothes, shoes, grocery, and health. We observed that over 20\% of the attribute values that are missing from the corresponding Amazon web pages can only be identified from the product images. We illustrate this with an intuitive example in Figure~\ref{fig:sample}. In general, we identified two main areas where the product images can be particularly useful:

\begin{itemize}
  \item \textbf{Additional information:} Important cues are
    sometimes absent in the product text descriptions. In Figure~\ref{fig:sample}(a), the brand of the product is
    \emph{``Lava''}, which is prominently displayed in the product
    image but not mentioned in the product title. In this case, OCR could perfectly
    recover the missing brand from the product image.
  
  \item \textbf{Cross validation:} The product title is likely to
    contain multiple possible values for one target attribute and the
    product image could help in disambiguating the correct value. In Figure~\ref{fig:sample}(b), for the
    attribute \emph{``Item Form''}, the product title contains the word
    \emph{``Stick''}, \emph{``Cream''} and \emph{``Powder''}. However, both the
    product shape and the word \emph{``Stick''} in the
    image strongly indicate the correct value should be
    \emph{``Stick''}.
\end{itemize}

Despite the potential, leveraging product images for attribute value extraction remains a difficult problem and faces three main challenges:
\begin{itemize}
\item {\bf C1: Cross-modality connections:} There are many intrinsic associations between product titles and images. An effective model needs to seamlessly and effectively make use of information from three modalities, including product images, texts in the images, and texts from product profiles.
  
\item  {\bf C2: Domain-specific expressions:}  The texts in the product images and
  product profiles are usually packed with certain phrases that are unique
  to a specific retail domain. For example, \emph{``cleaning ripples''}
  in the category of toilet paper is a special wavy pattern to help with
  cleaning. \emph{``free and clear''} in the category of detergent means that it is scent-free. In general, language models or word embeddings pre-trained on public corpora are unable to accurately capture and ground these domain-specific phrases.

\item {\bf C3: Multiple categories:} For the same attribute, there may be little overlap between the attribute values for different product categories. For example, the vocabulary for the attribute \emph{``size''} in T-shirts (\emph{i.e.}, small, median, large, x-large) is completely different from baby diapers
(\emph{i.e.}, newborn, 1, 2, 3, etc.).  Therefore,
  a model trained on one product category may generalize poorly to other categories. 
\end{itemize}

Existing solutions for multi-modal information extraction \cite{antol2015vqa,kim2018bilinear,lu2019vilbert,tan2019lxmert,singh2019towards} fall short in the e-commerce domain, as it cannot address challenges {\bf C2} and {\bf C3}.
On the other hand, text extraction solutions that manage to extract attribute values across multiple product categories~\cite{karamanolakis2020txtract} are text focused, and the techniques cannot easily be transferred to image information extraction. A comparison between these models are summarized in Table~\ref{tab:shortcomings}.
In this paper, we address the central question: {\em how can we perform multi-modal product attribute extraction across various product categories?}

\begin{table}[h]
  \small
  \caption{Comparison between Different Methods}
  \label{tab:shortcomings}
  \begin{tabular}{l|ccc|cc}
  \specialrule{0em}{-4pt}{0pt}
    \toprule
    \multirow{2}{*}{\textbf{Methods}}& \multicolumn{3}{c}{\textbf{C1}} & \textbf{C2} & \textbf{C3}\\\cline{2-6}
                                    & \textbf{Text} & \textbf{Image} & \textbf{OCR} & \textbf{Domain} & \textbf{Category}\\
    \midrule
    BAN\cite{kim2018bilinear} & $\surd$ & $\surd$ &  &  & \\
    LXMERT\cite{tan2019lxmert}  & $\surd$ & $\surd$ &  &  & \\
    LoRRA\cite{singh2019towards}  & $\surd$ & $\surd$ & $\surd$ &  & \\
    OpenTag\cite{zheng2018opentag} & $\surd$& &  & $\surd$ & \\
    TXtract\cite{karamanolakis2020txtract}& $\surd$& & & $\surd$ & $\surd$\\
    \midrule
    Ours & $\surd$ & $\surd$ & $\surd$ & $\surd$ & $\surd$\\
  \bottomrule
  \specialrule{0em}{0pt}{-3pt}
\end{tabular}
\end{table}

Inspired by recent progress on text visual question answering ({\em Text VQA)}~\cite{singh2019towards,hu2020iterative}, we address challenge {\bf C1} with a multi-modal sequence-to-sequence generation model. Our input comes from three folds: \emph{1)} the images, containing both OCR results to capture texts in the image, along with visual features generated through Faster RCNN~\cite{ren2016faster}, \emph{2)} the textual product profile, and \emph{3)} pre-defined possible values for the target attribute. The sequence-to-sequence generator allows the extraction output to expand beyond substrings mentioned in product profiles or images. We adopt a transformer-based model in order to incorporate cross-modal attention between texts and images. The pre-defined vocabulary, which can be retrieved from the training data, is used to address challenge {\bf C2}, by exposing the model to domain-specific target values.

We further extend our model to a cross-category extraction model to address challenge {\bf C3}. Towards this end, we utilize two different techniques:
First, instead of using a fixed pre-defined vocabulary, the model dynamically switches between different category-specific vocabularies based on the category information of the input product.  Second, we predict the product category as part of the decoding process, in a multi-task training setup. Instead of the common practice of introducing an auxiliary task as an additional training objective, we require our model to decode the product category as a part of the output. This implicit multi-task training fits naturally with our sequence-to-sequence model architecture.

We summarize the contributions of this work as follows:
\begin{itemize}
  \item We propose a transformer-based sequence-to-sequence model to extract product attributes jointly from 
    textual product profile, visual information,
    and texts in product images. To the best of our
    knowledge, this is the \emph{first} work for multi-modal product attribute extraction. 
  \item We extend our basic solution to a cross-product-category extraction model 
    by \emph{1)} equipping the model with an external dynamic vocabulary
    conditioned on product category and \emph{2)} multi-task training incorporated with our sequence-to-sequence model.
  \item We conduct extensive experiments to evaluate our
    solution on a dataset collected from a public e-commerce website across multiple product categories. Our approach
    consistently outperforms state-of-the-art solutions by $15\%$ on recall and $10\%$ on F1 metric.
\end{itemize}

The remainder of the paper is organized as follows: We first review
related work about attribute value extraction and visual
question answering in Section~\ref{related}. The problem formulation is given in
Section~\ref{definition}. In Section~\ref{sec:m4c} we describe
our model in the case of single product category. In Section~\ref{sec:pam}, we elaborate the
necessary components to make the model aware of different
product categories. Finally, we show the experimental results in
section \ref{experiment} and section \ref{conclusion} concludes
the paper.

\section{Related Work}\label{related}

\subsection{Value Extraction for Given Attributes}
There are many works on extracting product attributes from product
textual description. The open-tag model \cite{zheng2018opentag} used a
Bi-directional LSTM model followed by conditional random field
(CRF). Recent works focus on training a single model to perform value
extraction for multiple product categories or multiple attribute
types. Reference \cite{xu2019scaling} and \cite{wang2020learning}
proposed to include the attribute type in the model inputs so that
one model could extract value for different
attribute types based on this input. Reference \cite{wang2020learning} also introduced a
distilled masked language model loss to help the model generalize
better for attribute types or attribute values not seen in the training data. Reference
\cite{karamanolakis2020txtract} provided a solution for cross product category attribute value extraction. 
An auxiliary task of predicting
product category is used to improve the intermediate
representation. Product category is also represented by its Poincar\'e
embedding and used as inputs to compute attention in later
layers. 

Most of the works mentioned above formulate the attribute exaction
task as a sequence tagging problem, in which the output is the start
and end position of the attribute value in input text. This work formulates the task
as a sequence generation problem. The benefit is that the model can
produce the correct attribute value even if it is phrased differently 
in the input texts.

Utilizing image to extract product attributes has received attention
recently in, for example, \cite{zhu2020multimodal} where a gated
attention layer is introduced to combine information from product image and text. 
This work is different from \cite{zhu2020multimodal} in
that we utilize texts in product images which is shown to be
useful in product attributes extraction task.

\subsection{Visual Question Answering}

Our task is related to the visual
question answering (VQA) task in that both tasks study interaction between text and image modality. 
 Early works in VQA focus on the
design of the attention mechanism to merge information from image and text
modality, such as the bilinear attention in
\cite{kim2018bilinear}. The importance of words in the image to the VQA task was
first recognized in \cite{singh2019towards} which proposed a new
benchmark TextVQA dataset. Hu, et. al. \cite{hu2020iterative}
proposed to use transformer \cite{vaswani2017attention} to express a more general form of
attention between image, image objects, image texts and
questions. Recently \cite{yang2020tap} introduced pre-training tasks to
this model architecture that boosted the state of the  art of TextVQA
benchmark significantly.

Our work uses a multi-modality transformer architecture similar to the
one proposed in \cite{hu2020iterative}. This work is different from \cite{hu2020iterative} in that 
we address the challenges in product attribute value extraction which
do not exist in the text VQA task studied in \cite{hu2020iterative}. 
We achieve this by enabling the
model to query a dynamic external vocabulary which is conditioned on product
category and also introduce multi-task training so the model is aware
of product category.

\section{Problem Definition} \label{definition}
\begin{figure}[h]
  \centering
  \includegraphics[width=0.98\linewidth, trim = {0cm 0cm 0cm 0mm}, clip]{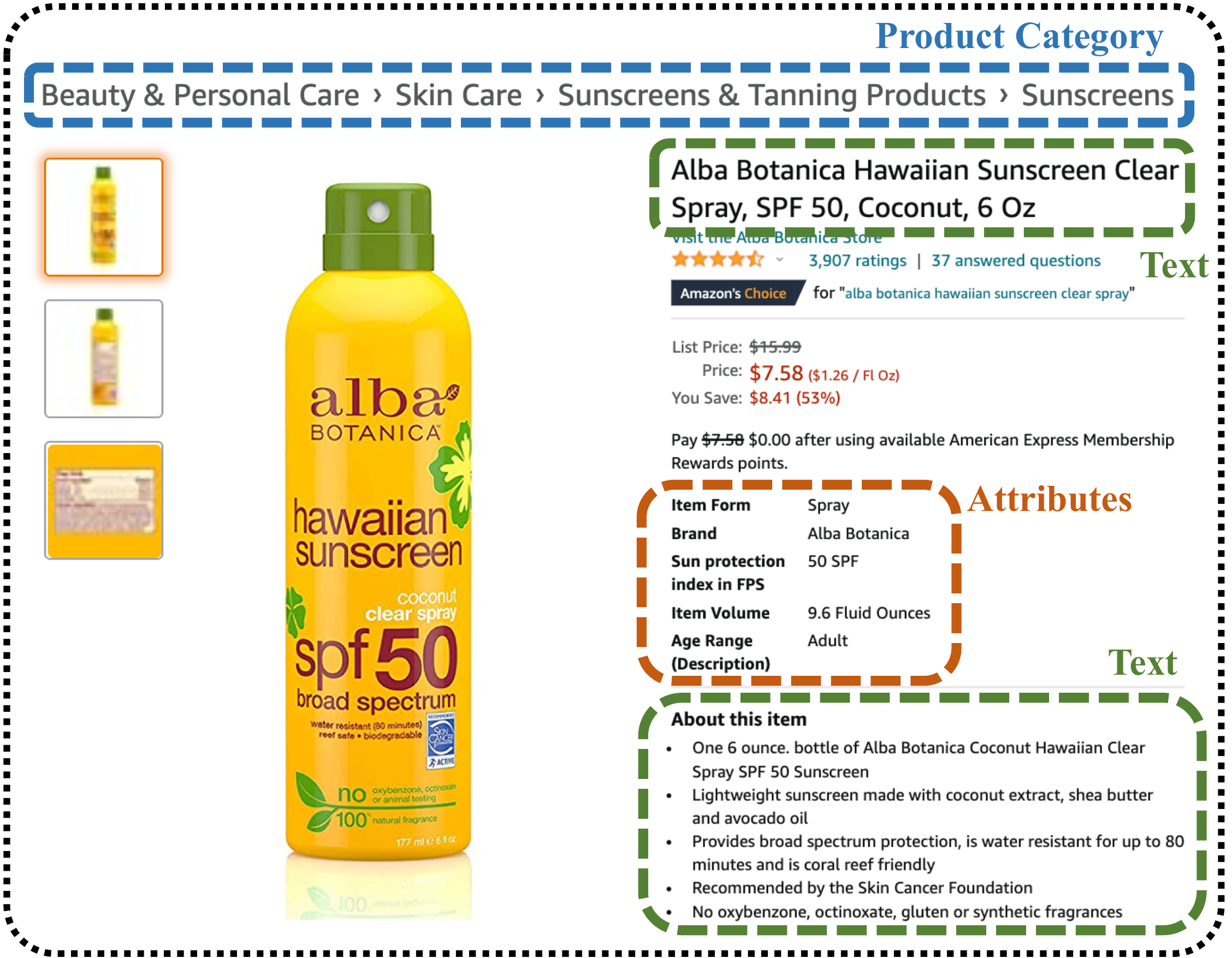}
  \caption{Example of product profiles.}
  \label{fig:product}
\end{figure}

A product profile displayed on an e-commerce website usually looks
like Figure~\ref{fig:product}. A navigation bar (top in
Figure~\ref{fig:product}) describes the category that the product
belongs to. Left side is product image and right side are product
texts, including a title and several bullet points. 

We consider products in a set of $N$ product categories $C = \{c_1, \cdots ,c_N\}$;
a category can be coffee, skincare, as shown in Table~\ref{tab:stat}.
We formally define the problem as follows. 

\smallskip
\noindent
{\bf Problem definition:} We take as input a target attribute $attr$ and a product with the following information:
\begin{enumerate}
\item a phrase describing the product category;
\item the text in the product
profile ({\em i.e.,} title, bullet points), denoted by a sequence of $M$ words $T = \{w_1^{text}, \cdots ,w_M^{text}\}$;
\item the product image \footnote{we use the first image shown on the
  website. For certain attributes such as nutrition information, later
  images such as nutrition label are used instead.}.
\end{enumerate}
Our goal is to predict the value for the target attribute $attr$. 
\smallskip

Figure~\ref{fig:product} displayed a few such attribute values
 for a sunscreen product.  Specifically, considering the
 target attribute \emph{``Item Form''} shown in Figure
 \ref{fig:product}, our objective is to extract the attribute value
 \emph{``Spray''}. If the target attribute is \emph{``Brand''}, the
 objective is to extract the attribute value \emph{'Alba Botanica'}.

\begin{figure*}[h]
  \centering
  \includegraphics[width=1\linewidth]{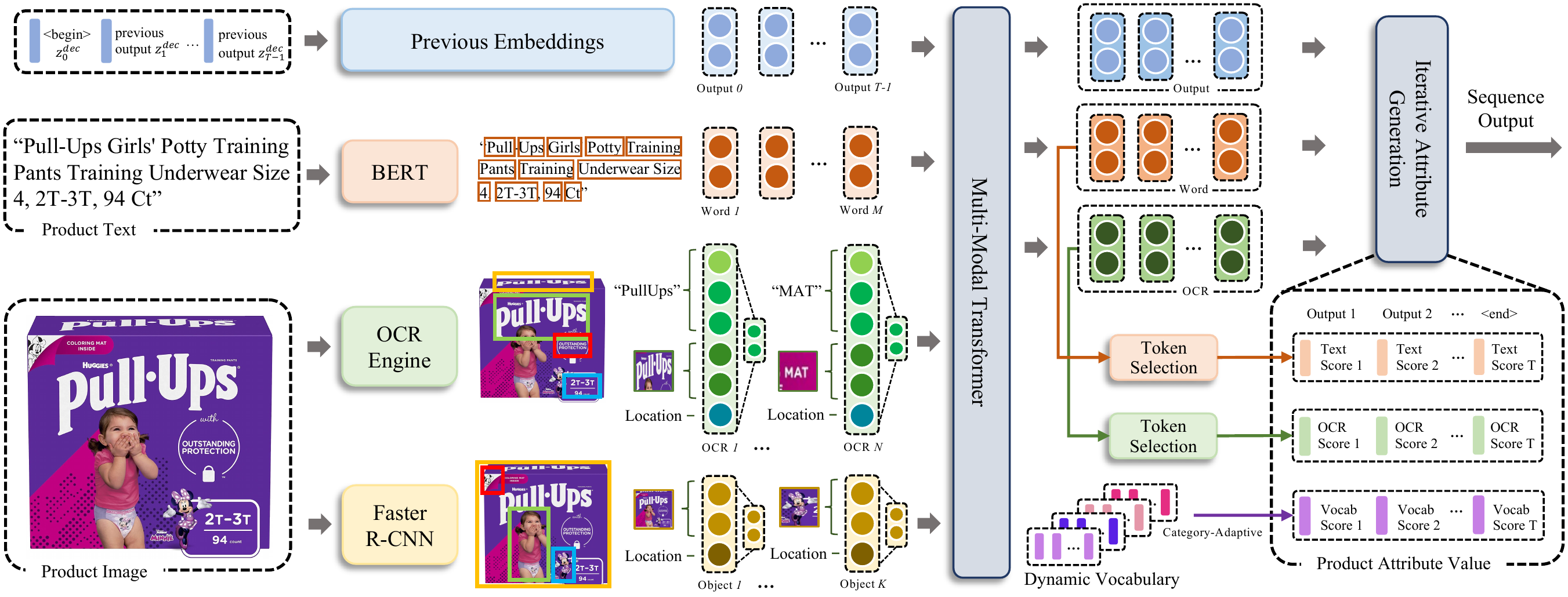}
  \caption{Overview of the proposed framework. 1) Input modalities: Product title and product image. 2) Token selection: Tokens could be selected from product title, OCR tokens identified in product image and a dynamic vocabulary conditioned on the product category. We consider edit distance between candidates and existing attribute values while selecting the next token. 3) Target sequence: We ask the decoder to first decode product category, and then decode attribute value.}
  \label{fig:overview}
\end{figure*}

 We next describe the model used to solve this problem. We first describe the model for the scenario of single product category (Section~\ref{sec:m4c}) and then describe its extension to the scenario of multiple categories (Section~\ref{sec:pam}). The model and its extension are illustrated in
 Figure~\ref{fig:overview}.

\section{Attribute Value Extraction for a Single Product Category} \label{sec:m4c}


\subsection{Overall Architecture: Sequence to Sequence generation model} As shown in Figure~\ref{fig:overview}, the overall model architecture is a sequence-to-sequence generation model. The encoder and decoder are implemented with one transformer, denoted as "Multi-Model Transformer" in Figure~\ref{fig:overview}, where attention masks are used to separate the encoder and decoder computations from each other internally. The input from the different modalities and the previously decoded tokens are each converted into vector representations, the details can be found in Section~\ref{sec:inputrep}. These representations are transformed into vectors of the same dimension, then concatenated as a single sequence of embeddings that is fed to the transformer. Therefore, an input position in each input modality is free to attend to other positions within the same modality, or positions from a different modality. The decoder operates recursively and outputs a vector representation $z_t^{dec}$ at the $t$th step. The decoder output is based on the  intermediate representations of the different encoder layers, along with the embeddings of the previously decoded tokens $z_{t-1}^{dec}$ at step $0 \cdots t-1$, denoted by "Previous Embeddings" in Figure~\ref{fig:overview}. A token selection module then chooses the token to output at step $t$ based on $z_t^{dec}$ from a candidate token set  (Section~\ref{sec:decoder}).

\subsection{Decoder Output Vocabulary and Token Selection}
\label{sec:decoder}
The traditional sequence-to-sequence generation model is known to
suffer from text degeneration \cite{holtzman2019curious}, in which the
decoder outputs repetitive word sequences that are not well formed
linguistically. To fix this problem, the output of the decoder in our model is
constrained to a set of candidate words. At the
$t$th decoding step, the decoder can return a token from the product text profile, the OCR engine, or from an external pre-defined
vocabulary. The external vocabulary is
composed of the words from the set of target attribute values, obtained from the training dataset. It is useful in the cases where \emph{1)}
the true attribute value is not mentioned in the product profile or image, \emph{2)}
the words in the image are not properly captured by the OCR algorithm, or \emph{3)} the true
attribute value is implied by the different inputs (from the product profile and images), but not explicitly mentioned. An example for the third case is predicting the target age of a ``hair clips for young girl'' product, where the target values include ``kid'', ``teenage'', or ``adult'' only. 

The output token at the $t$th decoding step is selected 
based on $z_t^{dec}$, the output vector
representation from the decoder at the $t$th step.  $z_t^{dec}$ is compared against the representation of every token in the
candidate set, and every token is assigned a score using a cosine-similarity formula. The three rows in the rectangle "Product Attribute Value" 
on the right bottom corner in Figure~\ref{fig:overview} illustrate the tokens
with the highest score within each of the three types of candidate tokens at T time steps. At any given time step, the token with the highest
score among all candidate tokens is chosen and output by the decoder. 

The scoring formulas depend on the source of the candidate token set, and are given by \eqref{eq:voc_score}, \eqref{eq:ocr_score} and \eqref{eq:text_score} for tokens from external vocabulary, OCR, and text respectively. 
\begin{align}
  y_{t,l} ^ {voc}&=(w_l^{voc})^T z_t^{dec}+b_l^{voc} \label{eq:voc_score}\\[2mm]
   y_{t,n} ^ {ocr}&=(W^{ocr}z_n^{ocr}+b^{ocr})^T(W^{dec}z_t^{dec}+b^{dec})\label{eq:ocr_score}\\[2mm]
  y_{t,m} ^ {text}&=(W^{text}z_m^{text}+b^{text})^T(W^{dec}z_t^{dec}+b^{dec}) \label{eq:text_score}
\end{align}

$y_{t,l}^{voc}$, $ y_{t,n} ^ {ocr}$ and $y_{t,m} ^ {text}$ are the
score for the $l$th word in the vocabulary of frequent answer words,
the $n$th OCR word, and the $m$th text token respectively in the $t$th
decoding step. The token in the external vocabulary is represented by
$w_l^{voc}$ which is a trainable vector. $z_n^{ocr}$ and $z_m^{text}$
are the output embeddings for the OCR token and the text token
respectively computed by the encoder. $b_l^{voc}$, $W^{ocr}$, $b^{ocr}$, $W^{text}$, $b^{text}$, $W^{dec}$, $b^{dec}$ represent the trainable parameters of
the linear transforms that align representations of candidate token
and $z_t^{dec}$.

\subsection{Representation of Inputs}
\label{sec:inputrep}

Previously decoded tokens are converted into vector representations and fed to the decoder. If the token is selected from product texts or OCR, it is represented by the output embedding from the encoder. If the token is selected from the external vocabulary, $w_l^{voc}$ in \eqref{eq:voc_score} is used as its representation.

The product profile texts are fed into the first three layers of a BERT model. The outputs of the pre-processing steps are then converted into three embeddings with the same dimension, as in \cite{hu2020iterative} (see Section~\ref{implementation} for implementation details).

The input image is pre-processed with object detection and OCR recognition. For object detection, the product image is fed into the Faster R-CNN object detection model \cite{ren2016faster}, which returns bounding boxes of detected objects, denoted as "Location" in Figure~\ref{fig:overview}, and fixed length vector representation extracted through RoI-Pooling among each detected region.

The OCR engine provides the detected texts, along with their bounding boxes. We also extract visual features over each OCR token's region using the same Faster R-CNN detector. We experimented with two different OCR solutions: \emph{1)} Public Amazon OCR API. \footnote{https://aws.amazon.com/rekognition} \emph{2)} Mask TextSpotter \cite{liao2020mask} which is more capable of detecting texts that are organized in unconventional shapes (such as a circle).

\section{PAM: Product-Category-Aware Multimodal Attribute Value Extraction Model}
\label{sec:pam}

The model described in the previous section performs reasonably well if trained and inferenced on a single product category. However, if it is trained and inferenced on multiple product categories, its precision and recall cannot outperform existing baselines and is sometimes even inferior. After analyzing the errors, we found that this happens because the output of the sequence generation model is not conditioned on product category. In this section we describe a model design that is aware of the product
category.

\subsection{Dynamic Vocabulary}

The vocabulary of target attribute values could contain very different words for different product categories. For example, it is unlikely for sunscreen brands to share common words with coffee brands except for words \emph{``company''} or \emph{``inc''}. Hence for each (product category, attribute type) pair, we pre-identify a specific vocabulary of frequent attribute value words, denoted by $V_{i,j}$ for the $i$th product category and $j$th attribute type. We also added the words that appear in the product category name to the vocabulary, we will clarify its role in Section~\ref{sec:multitask}. In addition, the word \emph{``unknown''} is added to the vocabulary, so the model can output \emph{``unknown''} when the input product profile does not contain a value for this attribute. The goal is to capture patterns for products where the attribute value is not usually conveyed. During the training process, the model will query the vocabulary $V_{i,j}$ according to the input product category $i$, which provides a more precise prior knowledge. Instead of training the representation from scratch as $w_l^{voc}$ in \eqref{eq:voc_score}, the vector representation for each word in $V_{i,j}$ is obtained with pre-trained fastText \cite{bojanowski2017enriching} embedding. This is because there are not enough training data in some product categories to compute a good word representation during the learning process.

\subsection{Domain Specific Output Token Selection}

The scoring function used for token selection,
\eqref{eq:ocr_score} and \eqref{eq:text_score}, are based on pre-trained embeddings. However, in our task the word could have domain-specific meaning. For example, \emph{``all''} and \emph{``off''} are considered stop words in English, but \emph{``all''}  is a popular brand of detergent and \emph{``off''} is a popular brand of insect repellent. The lack of domain-specific embedding hurts the accuracy of the token selection module. We therefore utilize the edit distance between the candidates and existing attribute values as a supplemental feature. We denote the edit distance based similarity ratio \footnote{The FuzzyWuzzy ratio implemented in https://github.com/seatgeek/fuzzywuzzy} for specific word token $w$ compared with the vocabulary $V_{i,j} = \{v_1,\cdots,v_L\}$ with $L$ words as:
\begin{align}
  f^e(w) = \max_{l \in L} \ \mbox{similarity\_ratio}(w, v_l)
\end{align}

With the decoded embedding $z_t^{dec}$ and the edit distance function $f^e$, we calculate the final score for candidates from the different modalities as follows, where \eqref{eq:voctoken}, \eqref{eq:ocrtoken}, and \eqref{eq:texttoken} are for tokens in the dynamic vocabulary $V_{i,j}$, OCR tokens, and tokens from product profile texts, respectively.
\begin{align}
y_{t,l}^{voc}&=(W^{voc}z_l^{voc}+b^{voc})^T(W^{dec}z_t^{dec}+b^{dec}) \label{eq:voctoken}\\[2mm]
y_{t,n}^{ocr}&=(W^{ocr}z_n^{ocr}+b^{ocr})^T(W^{dec}z_t^{dec}+b^{dec})+\lambda f^e(w_n^{ocr}) \label{eq:ocrtoken}\\[2mm]
y_{t,m}^{text}&=(W^{text}z_m^{text}+b^{text})^T(W^{dec}z_t^{dec}+b^{dec})+\lambda f^e(w_m^{text}) \label{eq:texttoken}
\end{align}

If $d$ denotes the dimension of the encoder's output embedding, then $W^{text}$, $W^{ocr}$ and $W^{dec}$ are $d\times d$ projection matrices, $W^{voc}$ is a $d \times 300$ matrix (300 is the dimension of the fastText embeddings). $z_m^{text}$, $z_n^{ocr}$ are the output embeddings for text tokens and OCR tokens computed by the encoder, respectively. $z_l^{voc}$  is the fastText embedding of the frequent vocabulary words. $b^{text}$, $b^{ocr}, $ $b^{voc}$, $b^{dec}$ are all $d$-dimensional bias vectors and $\lambda$ is the hyper-parameter to balance the score. Finally, the auto-regressive decoder will choose the candidate tokens with the highest score from the concatenated list $[y_{t,m}^{text},$ $\ y_{t,n}^{ocr},$ $\ y_{t,l}^{voc}]$ at each time step $t$.

\subsection{Multi-Task Training}
\label{sec:multitask}

We use the multi-task learning setup to incorporate the product categories in the overall model. We experiment with two methods of multi-task training. 

\subsubsection{Embed the Product Category in Target Sequence}

The first multi-task training method is based on
prefixing the target sequence the decoder is expected to generate during training with the product category name. For example, the target
sequence of product shown in Figure \ref{fig:product} would be
\emph{``sunscreen spray''} for attribute \emph{``Item Form''} and
\emph{``sunscreen alba botanica''} for attribute \emph{``Brand''}. The category name prefix serves as an
auxiliary task that encourages the model to learn the correlation
between the product category and attribute value. At inference time, the decoder output at the $t$th step depends on its previous outputs. But since the ground truth value of the product category is known, there is no need to depend on product category estimated by the decoder. We simply replace it with the true product category value. We have seen empirically that this modification improves the precision of the model. 

Let the target label during the $t$th decoding step be $[y_{t,m}^{text},$ $\ y_{t,n}^{ocr},$ $\ y_{t,l}^{voc}]$, which takes the value $1$ if the token from text, OCR, or external vocabulary is the correct token and $0$ otherwise. 
More than one token could have label $1$ if they are identical but from different sources. We use the multi label cross entropy loss between the target label list  $[y_{t,m}^{text},$ $\ y_{t,n}^{ocr},$ $\ y_{t,l}^{voc}]$  and the predicted score list $[\hat{y}_{t,m}^{text},$ $\ \hat{y}_{t,n}^{ocr},$ $\ \hat{y}_{t,l}^{voc}]$ given by \eqref{eq:voctoken}-\eqref{eq:texttoken}. 
The loss function hence contains two term: the loss contributed by the category name prefix, and the loss contributed by the attribute value in the target sequence:
\begin{align}
     & Loss = Loss_{\textrm{attribute value}} + \lambda_{cat} Loss_{\textrm{category name prefix}}    \label{eq:loss}
\end{align}
where $\lambda_{cat}$ is the tunable hyper-parameter to balance between these two losses.

\subsubsection{Auxiliary Task of Category Prediction}\label{classifier}
The target sequence method is by no means the only possible design of multi-task training. It is also possible to introduce a separate classifier $f^{cat}(z)$ to predict the product category. A specific classification token \Verb+<CLS>+ is inserted as the first entity in the input to the encoder. After concatenating and fusing with the other multimodal contexts in the transformer encoding layer, the enriched representation $z^{cls}$ corresponding to the classification token \Verb+<CLS>+ will be passed to the feed-forward neural network $f^{cat}(z)$ to predict the product category. 
\begin{equation}
    f^{cat}(z) = softmax(W^{cat}z^{cls}+b^{cat})
\end{equation}
where $W^{cat}$ and $b^{cat}$ are trainable parameters.

The training of the end-to-end model is jointly supervised by the sequence generation task and product category prediction tasks as described in \eqref{eq:overallloss}. 
\begin{align}
  Loss &= Loss_{\textrm{attribute value}} + \lambda_{cat} Loss_{cat}
  \label{eq:overallloss}
 \end{align}
 where $Loss_{cat}$ is the loss for product category prediction task.

\begin{table}
  \small
  \caption{Dataset Statistics}
  \label{tab:stat}
  \begin{tabular}{l|ccc}
    \toprule
    \textbf{Category} & \textbf{\# Samples} & \textbf{\# Attr1} & \textbf{\# Attr2}\\
    \midrule
    cereal & 3056 & 7 & 631\\
    dishwasher detergent & 741 & 8 & 114\\
    face shaping makeup & 6077 & 16 & 926\\
    fish & 2517 & 11 & 391\\
    herb & 6592 & 19 & 1220\\
    honey & 1526 & 20 & 472\\
    insect repellent & 994 & 20 & 373\\
    jerky & 3482 & 9 & 475\\
    sauce & 4218 & 10 & 878\\
    skin cleaning agent & 10904 & 22 & 3016\\
    skin foundation concealer & 8564 & 17 & 744\\
    sugar & 1438 & 10 & 347\\
    sunscreen & 5480 & 26 & 1295\\
    tea & 5719 & 14 & 1204\\
  \bottomrule
\end{tabular}
\end{table}

\section{Experiments Setup and Results}\label{experiment}
\subsection{Dataset}\label{data}
We evaluate our approach on 61,308 samples that cover 14 product categories. For each product category, we randomly collect the product texts, attribute values and images from the \Verb+amazon.com+ web pages. We split the dataset into 56,843 samples as training/validation set and 4,465 samples as held-out testing set. The attribute values shown on the web page are used as training label after basic pre processing, which handle the symbol and morphology issues. The attribute values labeled by annotators are used as benchmark testing label. Assuming the attribute type is applicable to the products, if the attribute value information can not be observed from the given product profile (text description, image, OCR tokens), we will assign \emph{``unknown''} as the corresponding value. In terms of the target attribute, we focus on two different types of attribute in the experiments. One of the criteria to determine the attribute type is the size of the value space. We consider attribute 1 \emph{``Item Form''} with around 20 values as an attribute with closed vocabulary, and attribute 2 \emph{``Brand''} with more than 100 values as an attribute with open vocabulary. Table \ref{tab:stat} summarizes the statistics of our dataset, where ``\# Samples'' denotes the number of samples, ``\# Attr1'' denotes the number of unique values for attribute 1 \emph{``Item Form''} and ``\# Attr2'' denotes the number of unique values for attribute 2 \emph{``Brand''}.

\subsection{Evaluation Metrics}
We use \emph{Precision}, \emph{Recall} and \emph{F1} score as the evaluation metrics. We compute \emph{Precision} (denoted as $P$) as percentage of ``match'' value generated by our framework; \emph{Recall} (denoted as $R$) as percentage of ground truth value retrieved by our framework; \emph{F1} score (denoted as $F1$) as harmonic mean of \emph{Precision} and \emph{Recall}. We determine whether the extraction result is a ``match'' using the exact match criteria, in which the full sequence of words are required to be correct.
\begin{table}
  \small
  \caption{Hyper-parameters Details}
  \label{tab:hyper}
  \begin{tabular}{l|c}
    \toprule
    \textbf{Hyper-parameters} & \textbf{Value}\\
    \midrule
    batch size &128\\
    iterations & 24,000\\
    optimizer & Adam\\
    base learning rate & $1e^{-4}$\\
    learning rate decay & 0.1\\
    learning rate decay steps & 15,000, 20,000\\
    max decoding steps  & 10\\
    transformer layers & 4\\
    transformer attention heads & 12\\
    embedding dimension & 768 \\
    lambda for edit distance & 0.05\\
  \bottomrule
\end{tabular}
\end{table}

\subsection{Implementation Details}\label{implementation}
The multi-modal encoder / decoder module is a 4-layers, 12-heads transformer which is initialized from scratch. More hyper parameters are illustrated in Table \ref{tab:hyper}. The dimension of the input representations are: \emph{1)} 100 product title tokens, \emph{2)} 10 image objects, \emph{3)} 100 OCR tokens. These representations are computed as following:

\subsubsection{Embeddings for Text Modality.}
The product text tokens are projected into sequence of vectors as the word-level embeddings using the first three layers of a BERT model, whose parameters are fine-tuned in training.

\subsubsection{Embeddings for Image Modality.}
The external object detection component is the Faster R-CNN model \cite{ren2016faster}, which has ResNet-101 \cite{he2016deep} as backbone and is pre-trained on the Visual Genome dataset \cite{krishna2017visual}. For each detected object, the Faster R-CNN model produces a fixed length vector which is a visual feature for the region of the object. In addition, the bounding box information of each object are also included as 4 coordinates. In order to combine and balance the energy of these two different types of features, projection fully-connected layer and the layer normalization 
are applied. During the training process, the final fully-connected layer of Faster R-CNN is fine-tuned. 

\subsubsection{Embeddings for OCR Modality.}
The OCR modality conveys both textual features (\emph{i.e.}
characters) and visual features (\emph{i.e.} color, font, appearance,
location of the tokens). For textual features, the OCR texts are
embedded using pre-trained fastText \cite{bojanowski2017enriching},
which could handle the out of vocabulary (OOV) words and morphology of
words properly. For robustness, additional character-level features are extracted by manually designed Pyramidal Histogram of Characters (PHOC) \cite{almazan2014word}. Similar to image embeddings, the pre-trained Faster R-CNN will represent the region of OCR tokens with its visual and coordinates features. Finally, linear projections are used
to cast these features into the same length, which are then added
together, passed through a layer normalization layer.

\subsection{Baselines}\label{baseline}
To evaluate our proposed framework, we choose the following models as baselines:
BiLSTM-CRF \cite{huang2015bidirectional},
OpenTag \cite{zheng2018opentag},
BUTD \cite{anderson2018bottom} and M4C \cite{hu2020iterative}. Our attribute value extraction task is highly related to the visual question answering tasks. Thus, among the four baselines, BiLSTM-CRF and OpenTag are attribute value extraction models, BUTD and M4C are visual question answering models. Some variants of our model and baselines are also included to make fair comparison on input sources. The details of baselines are listed below:
\begin{itemize}
    \item BiLSTM-CRF \cite{huang2015bidirectional}: the hidden states generated by the BiLSTM model are fed into the CRF as input features, the CRF will capture dependency between output tags. Text modality is used in this model.
    \item OpenTag \cite{zheng2018opentag}: on top of the BiLSTM-CRF, attention mechanism is introduced to highlight important information. Text modality is used in this model.
    \item BUTD \cite{anderson2018bottom}: Bottom-Up and Top-Down (BUTD) attention encodes question with GRU \cite{chung2015gated}, then attends to object region of interest (ROI) features to predict answer. Text and Image modalities are used in this model.
    \item M4C \cite{hu2020iterative}: cross-modality relationships are captured using multimodal transformer, the model will then generate the answer by iterative sequence decoding. Text, image and OCR modalities are used in this model. The answer could be selected from OCR tokens and frequent vocabulary.
    \item M4C full: to accommodate to the attribute value extraction task in e-commerce applications, extra input source of product title are added directly in the decoding process. Text, image and OCR modalities are used in this model. The answer could be selected from product title, OCR tokens and frequent vocabulary.
    \item PAM text-only: the text-only variant of our framework, image visual features and OCR tokens extracted from the product image are excluded from the input embeddings. Text modality is used in this model.
\end{itemize}
\begin{table}
  \small
  \caption{Comparison between the proposed framework PAM and different baselines}
  \label{tab:main}
  \begin{tabular}{c|lccc}
    \toprule
    \textbf{Attributes} & \textbf{Models} & \textbf{\emph{P(\%)}} & \textbf{\emph{R(\%)}}& \textbf{\emph{F1(\%)}}\\
    \midrule
    \multirow{7}{*}{Item Form}   & \multicolumn{1}{l}{BiLSTM-CRF} & \multicolumn{1}{l}{90.8} & \multicolumn{1}{l}{60.2} & \multicolumn{1}{l}{72.3}\\
                                 & \multicolumn{1}{l}{OpenTag} & \multicolumn{1}{l}{95.5} & \multicolumn{1}{l}{59.8} & \multicolumn{1}{l}{73.5}\\
                                 & \multicolumn{1}{l}{BUTD} & \multicolumn{1}{l}{83.3} & \multicolumn{1}{l}{53.7} & \multicolumn{1}{l}{65.3}\\
                                 & \multicolumn{1}{l}{M4C} & \multicolumn{1}{l}{89.4} & \multicolumn{1}{l}{52.6} & \multicolumn{1}{l}{66.2}\\
                                 & \multicolumn{1}{l}{M4C full} & \multicolumn{1}{l}{90.9} & \multicolumn{1}{l}{63.4} & \multicolumn{1}{l}{74.6}\\\cline{2-5}
                                 & \multicolumn{1}{l}{PAM (ours) text-only} & \multicolumn{1}{l}{94.5} & \multicolumn{1}{l}{60.1} & \multicolumn{1}{l}{73.4}\\
                                 & \multicolumn{1}{l}{PAM (ours)} & \multicolumn{1}{l}{91.3} & \multicolumn{1}{l}{75.3} & \multicolumn{1}{l}{82.5}\\
    \midrule
    \multirow{7}{*}{Brand}   & \multicolumn{1}{l}{BiLSTM-CRF} & \multicolumn{1}{l}{81.8} & \multicolumn{1}{l}{71.0} & \multicolumn{1}{l}{76.1}\\
                                 & \multicolumn{1}{l}{OpenTag} & \multicolumn{1}{l}{82.3} & \multicolumn{1}{l}{72.9} & \multicolumn{1}{l}{77.3}\\
                                 & \multicolumn{1}{l}{BUTD} & \multicolumn{1}{l}{79.7} & \multicolumn{1}{l}{62.6} & \multicolumn{1}{l}{70.1}\\
                                 & \multicolumn{1}{l}{M4C} & \multicolumn{1}{l}{72.0} & \multicolumn{1}{l}{67.8} & \multicolumn{1}{l}{69.8}\\
                                 & \multicolumn{1}{l}{M4C full} & \multicolumn{1}{l}{83.1} & \multicolumn{1}{l}{74.5} & \multicolumn{1}{l}{78.6}\\\cline{2-5}
                                 & \multicolumn{1}{l}{PAM (ours) text-only} & \multicolumn{1}{l}{81.2} & \multicolumn{1}{l}{78.4} & \multicolumn{1}{l}{79.8}\\
                                 & \multicolumn{1}{l}{PAM (ours)} & \multicolumn{1}{l}{86.6} & \multicolumn{1}{l}{83.5} & \multicolumn{1}{l}{85.1}\\
  \bottomrule
\end{tabular}
\end{table}

\subsection{Results and Discussion}
\subsubsection{Comparison between different model architectures}
We conduct experiments on the dataset described in section \ref{data}. We first show the performance comparisons between our approach, baselines and some variants on two different attributes \emph{``Item Form''} and \emph{``Brand''} in Table~\ref{tab:main}. As can be seen from these comparison results, PAM could consistently outperform the other baseline methods on Recall and F1 score. For example, for the \emph{``Item Form''} attribute, the Recall of PAM increases by 15\% compared with the text-only variant of PAM and increases by 22\% compared with the M4C model. 
For the \emph{``Brand''} attribute, the Recall of PAM increases by 5\% compared with the text-only variant of PAM and increases by 15\% compared with the M4C model. 
Note that PAM could achieve higher score on all metrics compared with the M4C full variant (full access to all modalities in the decoding process), which demonstrate the effectiveness of our task-specific designs on the framework.
There are two main reasons contribute to these improvements: \emph{1)} PAM utilizes rich information from naturally fused text, image and OCR modalities that could significantly improve Recall. These three modalities could help each other by providing important cues while information might be missing in specific modality. \emph{2)} PAM utilizes product category inputs in the decoding process. Attribute value is highly related to product category. By considering such crucial information, our model is able to learn enriched embeddings that could discriminate targeted attribute values from distracting values that belong to other product categories. 

\subsubsection{Usefulness of image, text and OCR inputs}
In order to quantify the impact of each modality, we further conduct ablation study on the input sources. We evaluate following variants of PAM:
\begin{itemize}
    \item PAM \emph{w/o} text is the variant that removes product texts modality from inputs.
    \item PAM \emph{w/o} image is the variant where features of detected objects are removed from inputs.
    \item PAM \emph{w/o} OCR is the variant that removes the OCR tokens from inputs.
\end{itemize}

From Table \ref{tab:modal} we can see that all the metrics on the attribute \emph{`Item Form'} degrade by removing any modality from the PAM framework, which demonstrates the necessity of combining all the modalities in our attribute extraction task. Closer inspection on the table shows that the text modality plays the most important role in this task. On the other hand, the image modality which represents the appearance of the product might be less effective compared to the other two modalities. The first possible reason is that the image could contain noisy information. In addition, similar shape of product might have different semantic meanings among various product categories. Finally, different attribute types also affect the performance, image modality could contributes more if the attribute type is related to color or obvious shape.

\begin{table}
  \small
  \caption{Usefulness of Image, text, OCR inputs}
  \vspace*{-2mm}
  \label{tab:modal}
  \begin{tabular}{l|ccc}
    \toprule
    \textbf{Models} & \textbf{\emph{P(\%)}} & \textbf{\emph{R(\%)}}& \textbf{\emph{F1(\%)}} \\
    \midrule
    PAM \emph{w/o} text & 79.9 & 63.4 & 70.7\\
    PAM \emph{w/o} image & 88.7 & 72.1 & 79.5\\ 
    PAM \emph{w/o} OCR & 82.0 & 69.4 & 75.1\\
    \midrule
    PAM  & 91.3 & 75.3 & 82.5\\
  \bottomrule
\end{tabular}
\end{table}

\begin{table}
  \small
  \caption{Impact of different components in the model}
  \vspace*{-2mm}
  \label{tab:category}
  \begin{tabular}{l|ccc}
    \toprule
    \textbf{Models} & \textbf{\emph{P(\%)}} & \textbf{\emph{R(\%)}}& \textbf{\emph{F1(\%)}}\\
    \midrule
    PAM \emph{w/o} target sequence & 88.5 & 72.9 & 80.0 \\
    PAM \emph{w/o} dynamic vocabulary & 89.1 & 69.5 & 78.1 \\
    \midrule
    PAM  & 91.3 & 75.3 & 82.5 \\
  \bottomrule
\end{tabular}
\end{table}

\subsubsection{Impact of different components in the model}
We also conduct experiments by removing each individual model design component from our framework to evaluate its effectiveness. The variants are listed below:
\begin{itemize}
    \item PAM \emph{w/o} target sequence is the variant that will generate attribute value without first generating the product category name. 
    \item PAM \emph{w/o} dynamic vocabulary is the variant that uses a
      large vocabulary of words shared by multiple categories instead
      of a dynamic vocabulary conditioned on product category.
\end{itemize}

Table \ref{tab:category} presents the extraction results on the \emph{``Item Form''} attribute. Without the category specific  vocabulary set, the model has to search on a much larger space for possible attribute values. The target sequence could enforce the category information via back-propagation loss. It is apparent from the results that these two category modules contribute to the final gains on Recall/F1 score.

\subsubsection{Comparison of Different Multi-Task Training Methods} \label{sec:multitaskresult}
The performance of the two multi-task training methods in Section \ref{sec:multitask} are compared in  
Table \ref{tab:multitask} for the \emph{`Item Form'} attribute. Overall, these two methods demonstrate
similar performance. 
\begin{table}
  \small
  \caption{Comparison of multi-task training}
  \vspace*{-2mm}
  \label{tab:multitask}
  \begin{tabular}{l|ccc}
    \toprule
    \textbf{Models} & \textbf{\emph{P(\%)}} & \textbf{\emph{R(\%)}}& \textbf{\emph{F1(\%)}} \\
    \midrule
    Auxiliary Task of Category Prediction & 90.8 & 75.4 & 82.3\\
    Embed Product Category in Target Sequence  & 91.3 & 75.3 & 82.5\\
  \bottomrule
\end{tabular}
\end{table}

\subsubsection{Comparison with Baseline on a Single Product Category}
Our approach is able to accommodate various product categories in one model. In order to verify such generalization ability on single category, we perform the category-level individual training tasks using the following baselines:
\begin{itemize}
    \item OpenTag\cite{zheng2018opentag}: the setting is the same as described in \ref{baseline}, except the training and evaluation are performed on single product category.
    \item Word Matching (WM): this is a brute-force matching baseline. \emph{1)} all the possible attribute values will be collected as attribute value dictionary $(\{\emph{``value''}: \emph{count}\})$, the count represents the popularity of corresponding attribute value; \emph{2)} manually exclude some distracting words like ``tea'' from the dictionary of the ``tea'' product category; \emph{3)} extract title tokens and OCR tokens for each sample; \emph{4)} compare the extracted tokens with attribute value dictionary in a popularity ascending sequence; \emph{5)} identify the attribute value if exact match is found in the attribute value dictionary. This baseline method does not require training, it is evaluated on the testing data directly.
\end{itemize}

\begin{table}
  \small
  \caption{Impact of OCR component on model performance}
  \vspace*{-2mm}
  \label{tab:ocrtoken}
  \begin{tabular}{l|cc}
    \toprule
    \textbf{OCR Detectors} & \textbf{average \# OCR tokens extracted} &  \textbf{\emph{F1(\%)}}\\
    \midrule
    Mask TextSpotter  & 13 & 71.1 \\
    Amazon Rekognition  & 42 & 80.3 \\
  \bottomrule
\end{tabular}
\end{table}

\begin{figure}
  \centering
  \vspace*{-6mm}
  \includegraphics[width=0.85\linewidth]{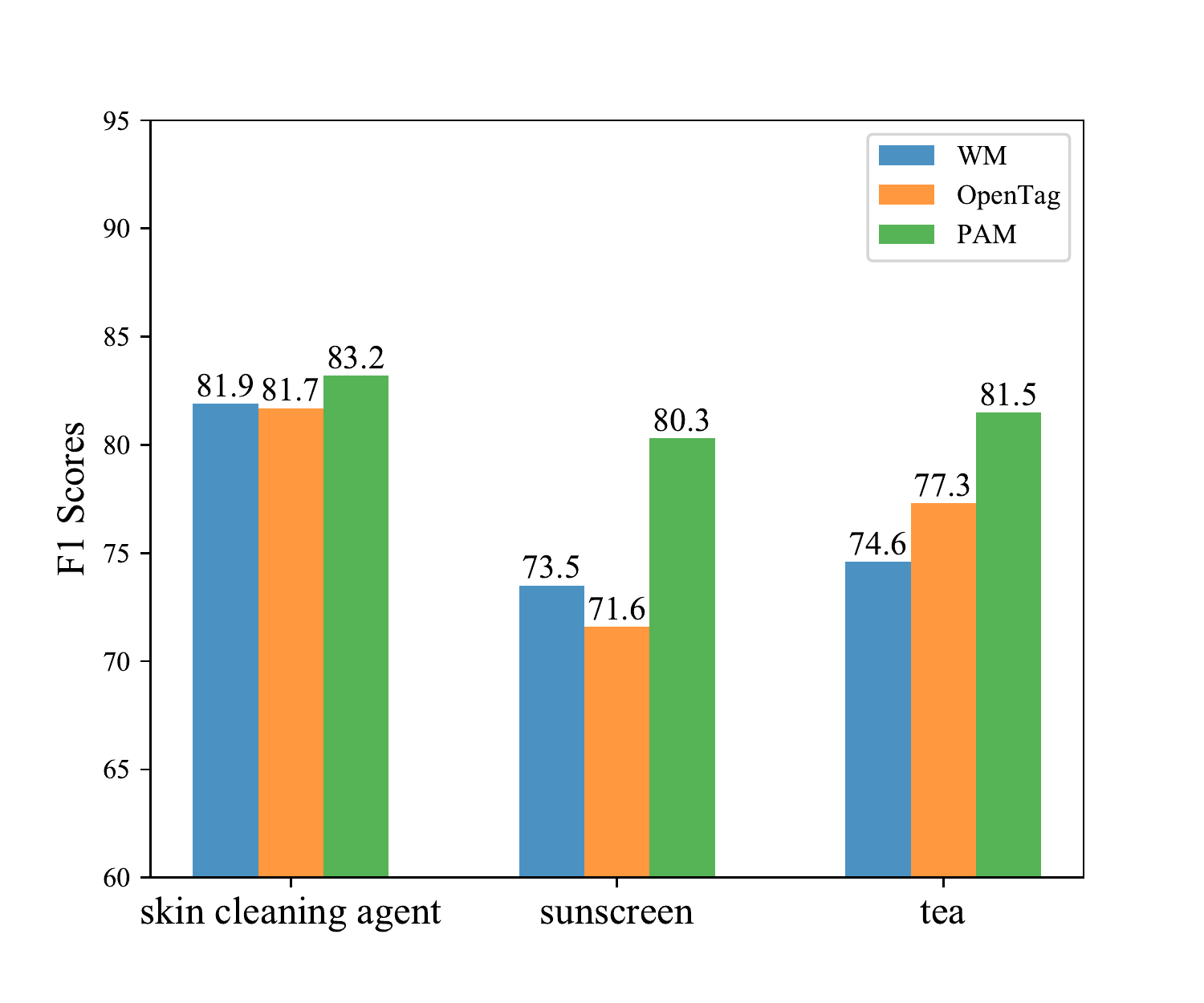}
  \vspace*{-6mm}
  \caption{Comparison between different methods on a single product category.}
  \vspace*{-2mm}
  \label{fig:single}
\end{figure}

For the purpose of performing category-level individual training, we
choose three categories that contain enough number of samples:
\emph{skin cleaning agent}, \emph{sunscreen} and \emph{tea}. Figure
\ref{fig:single} demonstrates the comparison of two baselines and our
model on single category. Our method consistently improves the
extraction metric. Although the WM baseline could also produce a good
F1 score for \emph{``skin cleaning agent''}, it requires manual efforts to
create a good list to exclude words that are impossible to appear in attribute
values, which is expensive to scale up to many product categories.

Under the single category settings, we also implement experiments to
evaluate the impact of the external OCR components on end-to-end
performance. As introduced in Section \ref{sec:m4c}, we use Amazon
Rekognition and Mask TextSpotter to extract OCR tokens from the
product image.  $F1(\%)$ is the extraction performance on attribute
\emph{Item Form} and category \emph{Sunscreen} of using corresponding
detectors. It can be seen from Table \ref{tab:ocrtoken} that
Rekognition is more suitable for our task. This is because Mask
TextSpotter is trained on a public dataset that is different from the
product image dataset. Therefore, Rekognition on average identifies
more OCR tokens and hence lead to better end-to-end $F1$ scores.

\vspace{-1mm}

\section{Conclusions}\label{conclusion}
To sum up, we explored a multimodal learning task that involves
textual, visual and image text collected from product profiles. We
presented a unified framework for the multimodal attribute value
extraction task in the e-commerce domain. Multimodal transformer based
encoder and decoder are used in the framework. The model is trained to
simultaneously predict product category and attribute value and its
output vocabulary is conditioned on the product category as well,
resulting in a model capable of extracting attributes across different
product categories.  Extensive experiments are implemented on a multi
categories/multi attributes dataset collected from public web
page. The experimental results demonstrate both the rich information
contained within the image/OCR modality and the effectiveness of our
product category aware multimodal framework.

For future works, pre-training task from \cite{yang2020tap} could be
useful in our attribute value extraction scenario. It is also valuable
to scale from 14 product categories to thousands of product categories
and model the complex tree structure of product categories properly
\cite{karamanolakis2020txtract}. The dynamic selection of vocabulary
in this framework could be incorporated into the training process as
in the RAG architecture \cite{lewis2020retrievalaugmented}.  Finally,
it is useful to design a model that extracts different attributes with
one model in which case the attribute generation order could be part
of the learning process too \cite{NEURIPS2019_1558417b}.

\section*{Acknowledgments}
\vspace{-1mm}
Rongmei Lin and Li Xiong are partially supported by NSF under CNS-1952192, IIS-1838200.
\vspace{-2mm}

\bibliographystyle{ACM-Reference-Format}
\bibliography{ref}

\end{document}